# Mining Trends of COVID-19 Vaccine Beliefs on Twitter with Lexical Embeddings

Harshita Chopra[1], Aniket Vashishtha[1], Ridam Pal[2], Ashima[2], Ananya Tyagi[2], Tavpritesh Sethi[2,3*]

1. Maharaja Surajmal Institute of Technology, GGSIPU, New Delhi, India
2. Indraprastha Institute of Information Technology Delhi, India
3. All India Institute of Medical Sciences, New Delhi, India

*tavpriteshsethi@iiitd.ac.in

## Abstract

**Background**

Social media plays a pivotal role in disseminating news globally and acts as a platform for people to express their opinions on various topics. A wide variety of views accompanies COVID-19 vaccination drives across the globe, often colored by emotions, which change along with rising cases, approval of vaccines, and multiple factors discussed online.

**Objective**

This study aims at analyzing the temporal evolution of different *Emotion* categories: Hesitation, Rage, Sorrow, Anticipation, Faith, and Contentment with *Influencing Factors*: Vaccine Rollout, Misinformation, Health Effects, and Inequities as lexical categories created from unsupervised word embeddings trained on the Tweets belonging to five countries with vital vaccine roll-out programs, namely, India, United States of America, Brazil, United Kingdom, and Australia.

**Methods**

We extracted a corpus of nearly 1.8 million Twitter posts related to COVID-19 vaccination and created two classes of lexical categories – Emotions and Influencing factors. Using cosine distance between word vectors to identify similar words, we expanded the vocabulary of each type. We tracked the longitudinal change in their strength over 11 months from June 2020 to April 2021 in each country. We used community detection algorithms to find modules in positive correlation networks.

**Results**

Our findings suggest that tweets expressing hesitancy towards vaccines contain the highest mentions of health-related effects in all countries. Our results indicated that the patterns of hesitancy were variable across geographies and can help us learn targeted interventions. We also observed a significant change in the linear trends of categories like hesitation and contentment before and after approval of vaccines. Negative emotions like rage and sorrow gained the highest importance in the alluvial diagram. They formed a significant module with all the influencing factors in April 2021, when India observed the second wave of COVID-19 cases. The relationship between Emotions and Influencing Factors was found to be variable across the countries.

**Conclusion**
By extracting and visualizing these, we propose that such a framework may help guide the design of effective vaccine campaigns and be used by policymakers to model vaccine uptake.

**Keywords**
Vaccination; Word Embeddings; Hesitancy; COVID-19; Twitter

## 1. Introduction

The unprecedented spread of COVID-19 has created massive turmoil in public health around the world. The development of vaccines has played a pivotal role in eradicating and mitigating significant outbreaks of infectious diseases like smallpox, tuberculosis, measles, and more such contagious diseases [1]. Major pharmaceutical companies located across the globe are in the phase of developing vaccines, with only a handful of them being authorized for clinical trials [2,3]. As the distribution of vaccines and associated campaigns expand, people continue to express their opinions and personal incidents on social media platforms.

Social media plays a decisive role in propagating information, leading to the emergence of varying perceptions related to the pandemic [4]. During the initial phase of national lockdown in several countries, Twitter had reported an increase of 24% of daily active users due to the increased usage of social media, the highest year-over-year growth rate reported by the company to date [5]. The COVID-19 pandemic has been studied in multidisciplinary aspects, and the analysis of Twitter posts remains a widely explored area in public health research [6,7,8] primarily because of the rapidly evolving nature of the content. Over the last decade, researchers have used multiple methods such as sentiment classification [9], social network analysis [10], and topic identification [11] to study the presence of pro-vaccine and anti-vaccine communities on social media. It is observed that vaccine uptake is affected by multiple factors, including rising adverse effect reportings, socioeconomic inequities, and quantitative allocation [12]. In addition, the spread of misinformation online has been a concerning issue, and prior survey-based studies suggest that it is linked with vaccine hesitancy and effects on public health [13,14]. On the other hand, certain marginalized groups continue to face inaccessibility to vaccines [15].

This paper presents a temporal and demographic analysis of lexical categories mined from Twitter conversations around vaccines. We further subdivided these categories into two subtypes: Emotions and their Influencing Factors. We examined the relationships between Emotions such as Hesitancy, Anxiety, Contentment, Sorrow, and Faith with Influencing Factors such as conspiracy theories around vaccines, social inequities, and health effects using unsupervised word embeddings trained on the curated corpus of tweets for eleven months. Further, we created correlation-based networks of these categories and performed clustering using the Infomap algorithm. The alluvial diagrams generated by these networks demonstrate the flow of importance of each factor from one month to another. We performed a granular analysis of the temporal-based trends of various outlooks towards COVID-19 vaccine activities. We analyzed their correlation with prominent factors for five countries, India, USA, Brazil, UK, and Australia, located on five different continents to demonstrate the comparative results among them.

While there has been various research work recently to analyze vaccine hesitancy or sentiment analysis for determining the overall general perception among people towards COVID-19 vaccines, our work provides a much more detailed insight into the variety of outlook people had towards the emergence of continuous vaccine updates and the possible reasons it can be correlated with. Major analysis work has been done on survey data conducted on a specific region, or a cohort of the population has to understand people's opinions toward vaccine uptake or resistance. Still, we have worked on a large corpus of tweets (more than 1.8 million) from different countries. As the meteoric rise in the use of social media has become a substantial influencing source in formulating different perceptions in millions of users, therefore, working on such a data source helps in getting a broader and better sense of various factors that might be associated with fuelling vaccine resistance. We have also analyzed our findings with vaccine developments and news in each country during the specific time periods to support our results.

## 2. Methods

**2.1 Design and Dataset.**
We performed an observational study by curating a longitudinal dataset by scraping more than 1.8 million tweets using the Snscrape library [16], starting from June 2020 to April 2021. The query used for extracting the tweets was created using an 'OR' combination of hashtags and words related to vaccines, and the name of the vaccines administered in the respective countries. Detailed queries for each country are mentioned in Table 1.

| Country | Query | No. of Tweets |
|---------|-------|---------------|
| USA | (General keywords) OR (moderna OR pfizer OR biontech OR astrazeneca OR inovio OR novavax OR #pfizerbiontech) | 1121216 |
| UK | (General keywords) OR (pfizer OR biontech OR oxfordvaccine OR astrazeneca OR moderna OR #pfizerbiontech) | 432271 |

| India | (General keywords) OR (covishield OR covaxin) | 229127 |
| Australia | (General keywords) OR (pfizer OR biontech OR oxfordvaccine OR astrazeneca OR moderna OR novavax OR #pfizerbiontech) | 50224 |
| Brazil | (General keywords) OR (coronavac OR Sinovac OR AstraZeneca OR Pfizer OR BioNTech OR #pfizerbiontech OR oxfordvaccine) | 17608 |

*Table 1: Queries used for scraping tweets from each country and number of tweets used after preprocessing. General keywords: (vaccine OR vaccination OR vaccinate OR covax OR #covidvaccine OR #coronavaccine OR #covidvaccination)*

Pre-processing of tweets was carried out on lowercase converted text by removing white spaces, punctuation, hashtags, mentions, digits, stopwords, URLs, and HTML characters. The verbs present in the text were lemmatized using WordNetLemmatizer from the NLTK package [17]. Duplicate tweets were removed based on identical username, time, and location. Figure 1 illustrates an abstract view of the study design.

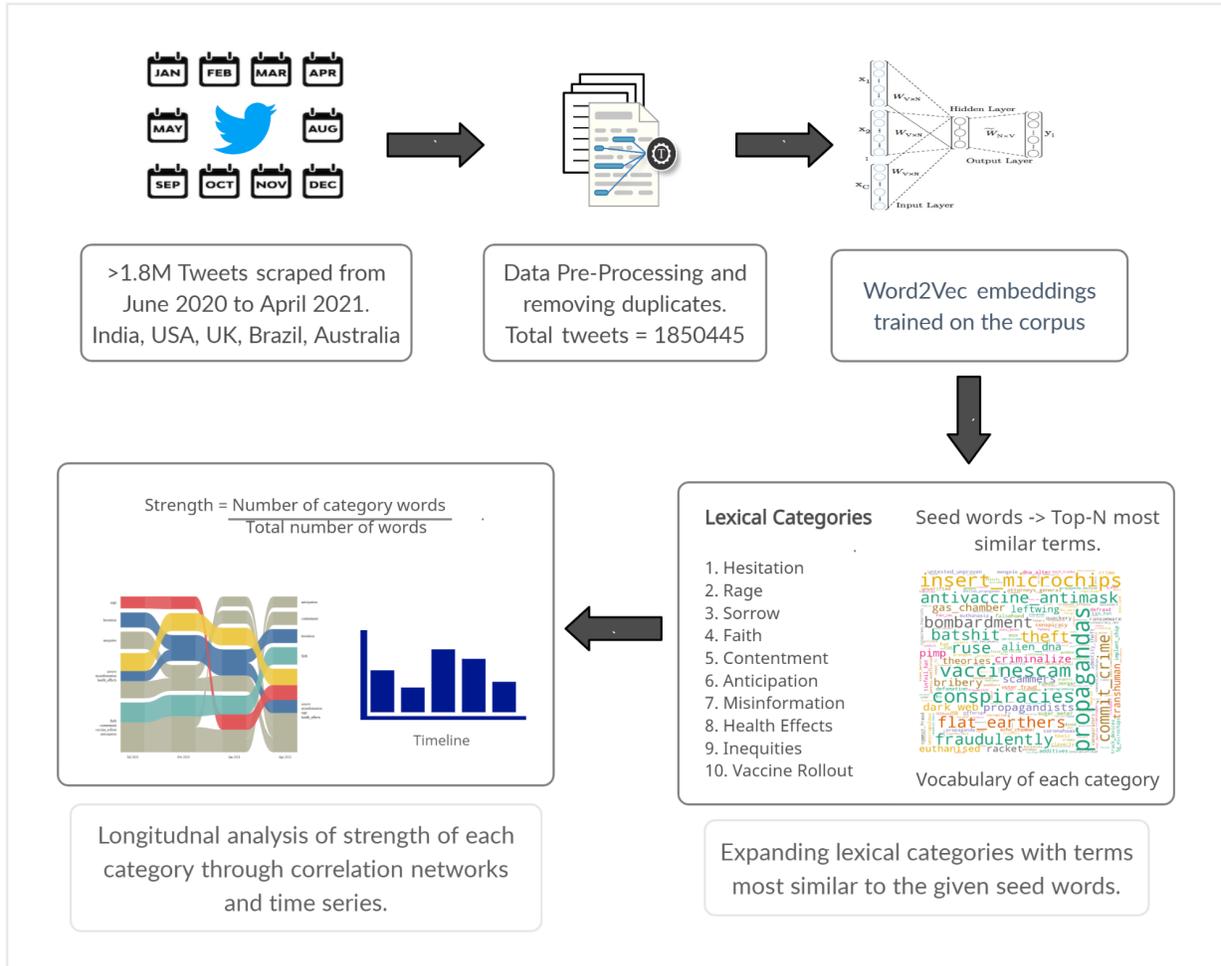

*Figure 1: Overview of the pipeline followed to create and analyze the strength of lexical categories.*

**2.2 Curating Categories using Unsupervised Word Embeddings.**

We created ten lexical categories for a psychometric evaluation of the tweet content in an approach similar to Empath [18]. The categories formed can be broken down into two classes: 'Emotions' and 'Influencing Factors.' Emotions consist of the affective processes that help us understand how reactions, feelings, thoughts, and behavior of people evolve in a given situation. We selected six COVID-19 related emotions, namely, Hesitation, Anxiety, Sorrow, Faith, Contentment, and Urgency, along with their putative influencing factors such as Misinformation, Vaccine Rollout, Inequities, and Health Effects in contrast to the COVID-19 vaccines. We specified a set of seed words corresponding to these categories, as shown in Table 1.

We trained a low dimensional representation ($d = 100$) as word embeddings for the unigrams and frequently occurring bigrams (co-occurring at least five times with the bigram scoring function [19] greater than a threshold of 50) present in our corpus using the skip-gram algorithm of the

Word2Vec model [20] with a sliding window size of five. We defined lexical categories as sets of words most similar to the assigned seed words. Each seed word, ensured to be present in the model's vocabulary, was mapped to a word vector. We used cosine similarity to measure proximity to find the top N(=50) words in the nearby vector space. Following this approach, $k$ seed words were expanded to a list of maximum $k \times N$ words. A category was defined as the union set of seed words and their closest similar words (Table 2). Seed words used for the Health Effects category were taken from the adverse events mentioned in the VAERS database [21], which occurred in our dataset's vocabulary. The resulting set of words in each lexical category were manually verified.

|  | Category | Description | Seed words |
|---|---|---|---|
| **Emotions** | | | |
| 1. | Hesitation | Skeptic attitude and reluctance towards being vaccinated due to multiple negative factors affecting an individual's opinions. | anxious, nervous, fear, consequences, uncertain, hesitation, suspicion, harm |
| 2. | Sorrow | Dissatisfaction and disapproval towards the different phases of COVID-19 vaccine production and distribution. | sad, hopeless, worst, disappointment, setback |
| 3. | Faith | Signifies strong belief and confidence in vaccines along with optimistic behavior towards the success of vaccines. | faith, optimism, vaccines work, assurance, grateful |
| 4. | Contentment | Signifies a state of happiness, appreciation, and acceptance of the COVID-19 vaccines. | satisfy, glad, proud, gratitude, great, joy |
| 5. | Anticipation | State of urgent demand and necessity of vaccines. | anticipate, urgently, priority, quick, await |
| 6. | Rage | Anger or aggression is associated with conflict arising from a particular situation. | angry, annoyance, hate, mad, pathetic |
| **Influencing Factors** | | | |
| 7. | Misinformation | Propagation of false information such as misinterpreted agendas and conceiving vaccines as conspiracy or scam. | propaganda, conspiracy, fraud, fake, poison |

| 8. | Vaccine Rollout | Availability and distribution of vaccines through campaigns and mass vaccination drives. | vaccinate, distribution, supply, mass, dose, vaccination drive |
| --- | --- | --- | --- |
| 9. | Inequities | Socioeconomic disparities are based on societal norms such as caste, race, religion, etc. | socioeconomic, deprive, racial_injustice, racism, underrepresented |
| 10. | Health Effects | Mentions of health-related adverse events caused by or affected by vaccines, including diseases, symptoms, and pre-existing conditions. | From the VAERS database. Eg. headache, fatigue, inflammation, etc. |

*Table 2: Curated categories: Emotions and Influencing Factors, their description, and seed words*

**2.3 Temporal Analysis of Lexical Categories.**
To measure each category's strength in a given text, we used the word count approach, similar to Empath [18] and other lexicon-based tools like LIWC [22]. To obtain an unbiased value that is independent of the length of text, we divided the frequency by the total number of words using the following formula,

$$Strength\ of\ Category\ (S) = \frac{Number\ of\ occurrences\ of\ category\ words\ in\ text}{Total\ number\ of\ words\ in\ text}$$

We appended the pre-processed text of all tweets monthly to calculate the strength. The time series of the strength of Emotion categories and Influencing Factors was helpful in analyzing the evolution of perceptions and opinions expressed by the public and how they vary with crucial timestamps like the news of the country's first vaccine approval.

**Analysis of Change Before and After Approval.**
To understand the variation of emotions amongst social media users in the aftermath of approval of vaccines, we conducted a Before-After change analysis for each lexical category based on the date when the country's government approved the first COVID-19 vaccine.
We created a day-wise time series of the strength of each category from June 2020 to April 2021 and smoothened it using the Moving Average algorithm. The linear nature of the trend was captured using an ordinary linear regression model fit on the strength of a category in the two time periods preceding and succeeding the approval date. To calculate the significance of the change, we used the z-test to compare the regression coefficients [23].

$$z = \frac{b_1 - b_2}{\sqrt{SE_{b_1}^2 + SE_{b_2}^2}}$$

Where $b_1$ and $b_2$ denote the slopes and $SE_{b_1}$ and $SE_{b_1}$ are the standard errors of the regression lines and before and after the approval, respectively.

To understand the Influencing factors co-occurring with Hesitation, we resampled the tweets having a positive strength of Hesitation (n=1000) and calculated the percentage of tweets that also have a positive strength of Anticipation, Rage, Misinformation, Health Effects, and Inequities. The resampling was repeated for 100 iterations, and its mean and standard errors were plotted (Figure 4). The percentage of tweets from each of these categories changed before and after the approval was recorded and tested for significance.

**Longitudinal Correlation-based Networks.**
The correlation between any two categories represents the degree to which they are linearly related. Daily strengths were calculated for each category followed by pairwise Pearson correlation [24]. Weighted networks of categories (nodes) and edge-strengths (correlations coefficients) were constructed for evaluating the positive associations among classes ($\rho \geq 0$). Community detection on these networks was carried out using the Infomap algorithm [25], and the dynamic change in these associations was visualized as an alluvial diagram [26].

## 3. Results

### 3.1 Analysis of Lexical Categories

Unsupervised word embeddings capture the context of words in the latent space based on their distribution and patterns of co-occurrence [27]. Given the noisy nature of social media data, it becomes difficult to implement a pre-defined lexicon-based approach with appropriate semantic inclusion. In this paper, we used unsupervised word embeddings trained on our corpus of tweets for finding the words most similar to a given set of seed words, hence expanding the vocabulary of a lexical category. Figure 2 shows the words belonging to the category of Misinformation and Hesitation. Some of the words most similar to 'conspiracy' were found to be 'implant_microchips' ($\cos\theta = 0.844$), 'qanon_conspiracy' ($\cos\theta = 0.820$), 'tinfoil_hat' ($\cos\theta = 0.808$) and 'echo_chamber' ($\cos\theta = 0.806$). These terms denote how people link vaccines to unconventional concepts and propaganda. The lexical category of Hesitation represents words such as 'skeptical,' 'disillusionment,' 'needle phobic,' 'dissonance,' and 'consequence' which demonstrate the uncertainty and doubt regarding vaccines and their effects.

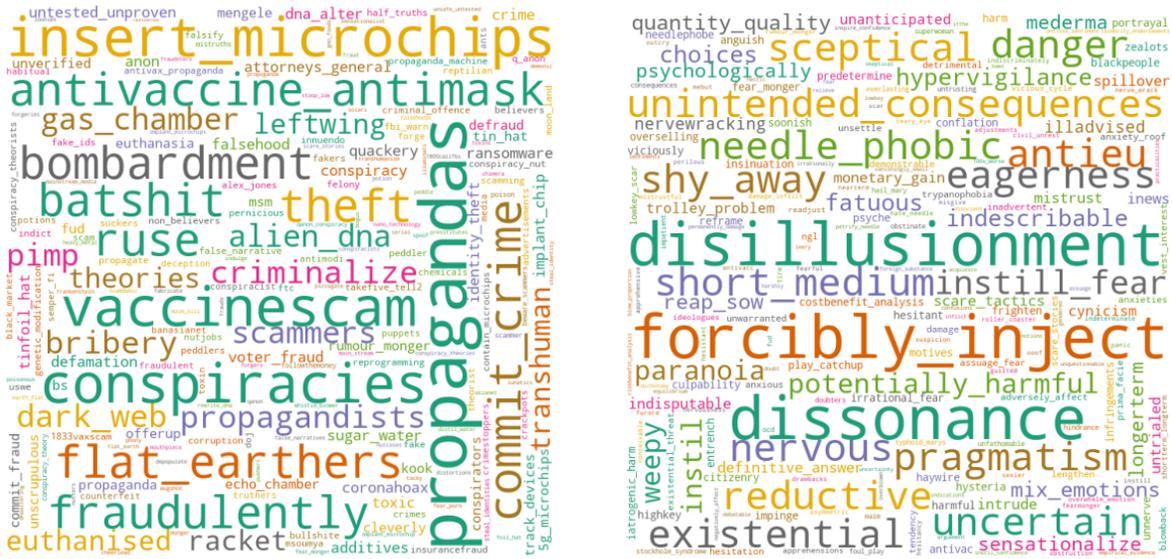

*Figure 2: (a) Words belonging to the lexical category of Misinformation. (b) Words belonging to the lexical category of Hesitation. These words represent the vocabulary expanded from the seed words of the respective categories.*

### 3.2 Change of Trends Before and After Approval.

The difference in slopes of linear trends of *Before* and *After* periods for each category demonstrate two significant inferences: the magnitude of change and the direction of change. Figure 3(a) shows the trends of Hesitation in India. A significant change in the direction of the slope is observed, which depicts a decrease in its strength after the approval (p<.001). There was a significant increase (p<.001) in the magnitude of tweets expressing contentment during the vaccination phase in the USA, as shown in Figure 3(b).

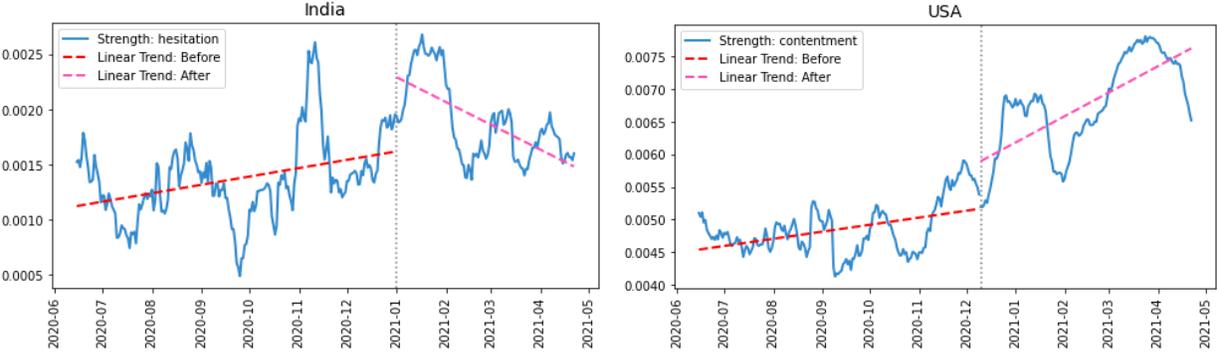

*Figure 3: Linear variation in the strength of (a) Hesitation in India and (b) Contentment in the USA. The dotted line represents the date of approval.*

The percentage of tweets belonging to different categories was analyzed from the sample of tweets before and after the approval of vaccines in each country. Figure 4(a) shows that Faith and

Contentment, both, were significantly higher (p<.001) before the approval of the first vaccine in India on January 01, 2021 [28]. The factors co-occurring with hesitation were analyzed by calculating the percentage of tweets of five other categories (Figure 4(b)). Our findings suggest that mentions of health effects contributed the most in tweets with a positive hesitation score. Rage and discussions on misinformation became significantly higher (p<.001) in the vaccination phase in India, while an opposite trend was observed in the USA after approval on December 10, 2020 [29]. A similar analysis for the UK, Brazil, and Australia is shown in Supplementary Figure 1.

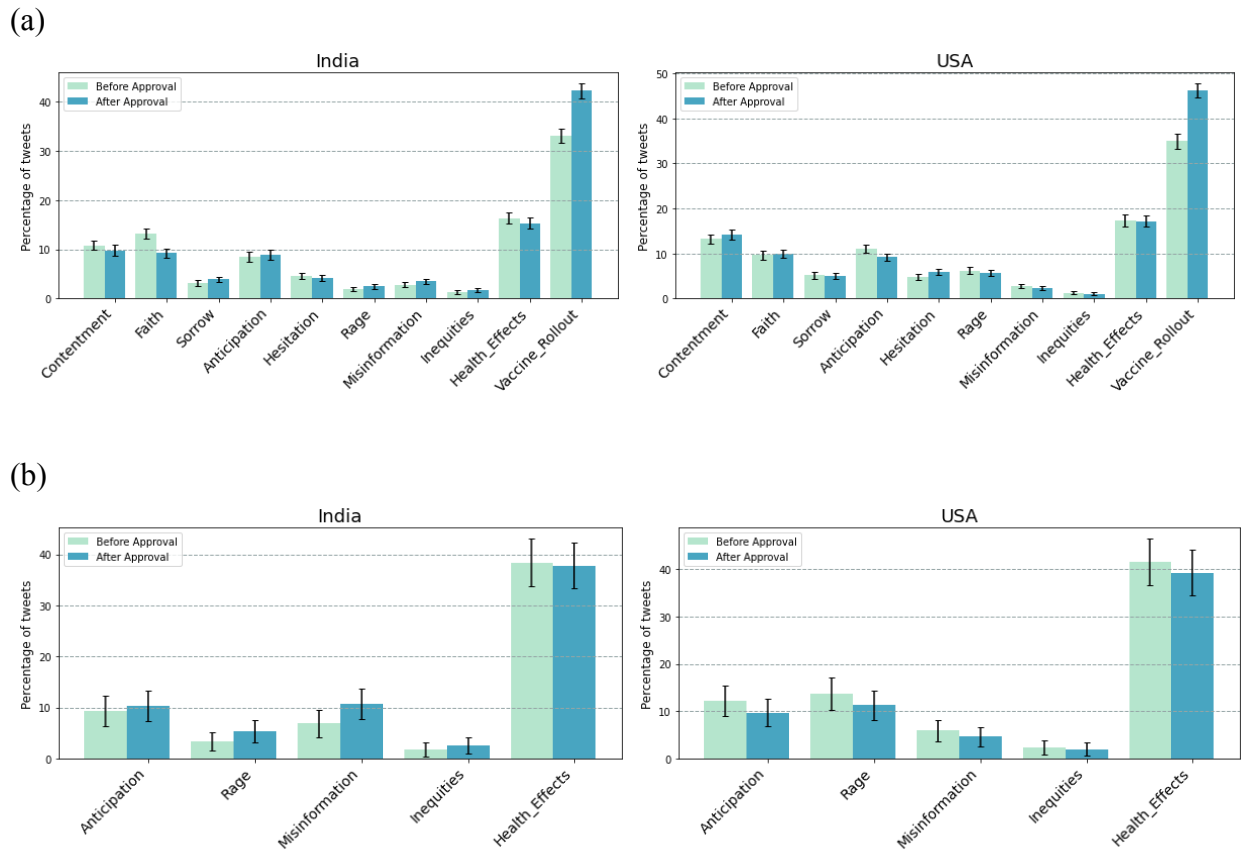

*Figure 4: (a) Percentage of tweets having a positive strength in each lexical category before and after approval of COVID-19 vaccine in India (Jan 01, 2021) and USA (Dec 10, 2020). (b) Percentage of Anticipation, Rage, Misinformation, Inequities, and Health Effects in positive 'hesitancy' tweets from India and USA.*

### 3.3 Longitudinal Analysis using Alluvial diagram

Inferences from the alluvial diagrams (Figure 5) based on Infomap clustering on correlation networks demonstrated that all the Influencing Factors, i.e., Misinformation, Health Effects, Inequities and Vaccine Rollout formed a primary module with emotions of sorrow and rage which gained the highest PageRank in April 2021, the time when India saw the second wave of COVID-19 cases while the vaccine rollout continued. This articulates the stern sentiment of disappointment due to rising issues and the non-availability of vaccines for people under the age group of 45 years. It also had a high correlation with tweets mentioning the spread of misinformation. Faith, contentment, and anticipation, which were found to be highly associated in the early months of July and October 2020, were found to be relatively less important and unrelated in April 2021.

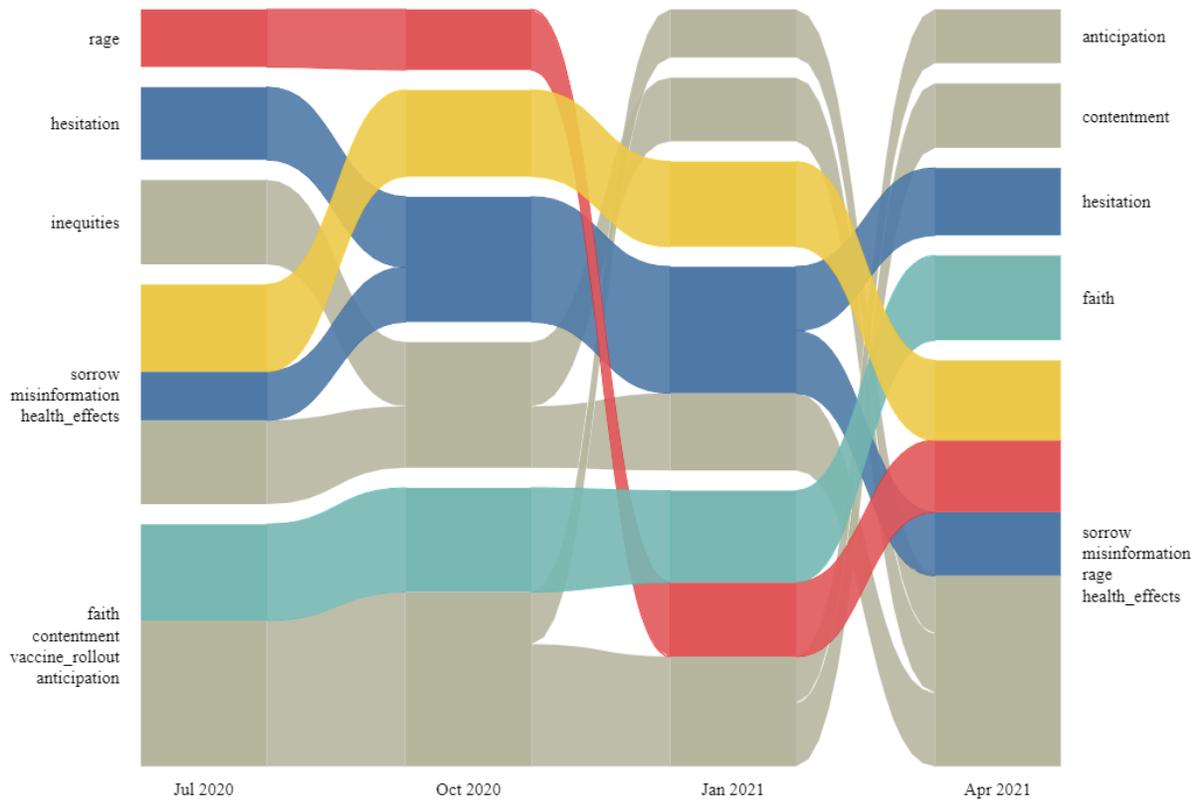

*Figure 5: Alluvial diagram for correlation-based networks showing the evolution of categories from July 2020 to April 2021, on an interval of three months, in India.*

On the contrary, lexical categories representing positive sentiment in the USA evolved to a significant module. Faith, contentment, and anticipation towards the vaccine were found to have a positive correlation with each other (Figure 6). Hesitation was the emotion influenced by mentions of health effects and inequities, whereas rage, sorrow, and misinformation were seen as less central factors in the USA.

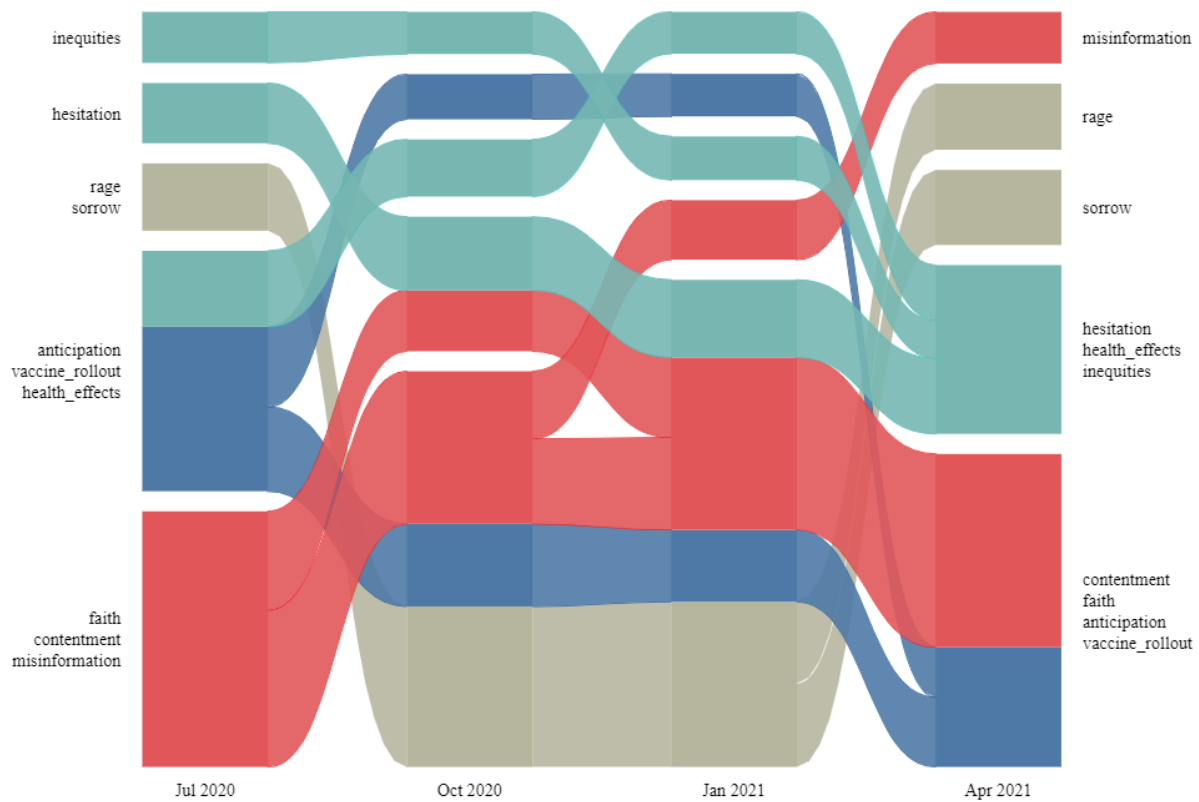

*Figure 6: Alluvial diagram for correlation-based networks showing the evolution of categories from July 2020 to April 2021, on an interval of three months, in the USA.*

Analysis of the temporal trend of vaccine rollout with hesitation and misinformation in five countries has been depicted in Figure 7. Updates regarding vaccinations started increasing near the end of 2020, which led to changing trends of hesitation expressed on Twitter. A notable inference from the line plots depicts that hesitation started rising from the beginning of 2021 when primary vaccination drives were initiated. In addition to this, rage is highly expressed in the tweets from the USA, while mentions of misinformation-related terms have been in more significant proportion in India and the UK. Lexical categories of hesitation and rage were found to have similar trends, suggesting a tentative association between the two. This is conclusive evidence that validates the spread of misinformation over social media.

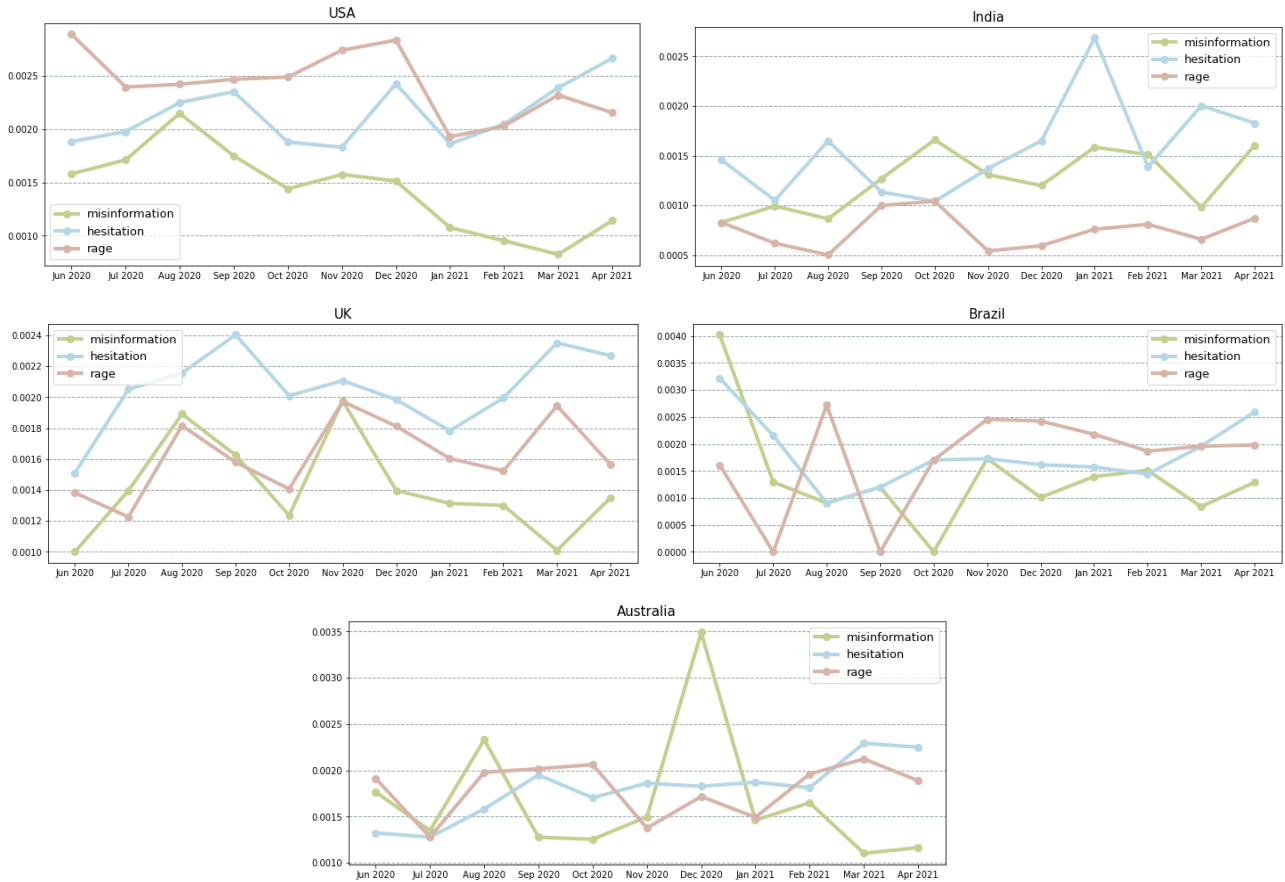

*Figure 7: Comparing the temporal flow of strength of two categories (vaccine rollout, misinformation, and hesitation) for five countries USA, India, UK, Brazil, and Australia.*

## 4. Discussion

The rise in social media platforms, such as Twitter has resulted in a valuable source to understand temporal variation in multiple affective and social categories. Influencing factors represented by word embedding-based lexical categories, namely, Misinformation, Vaccine Rollout, Inequities, and Health Effects, significantly assisted in studying the public perceptions towards emerging vaccine updates from initial approvals to rollout and administration.

### 4.1 Interpretation of Results

Widespread misinformation getting articulated through social media creates panic among the users [30]. The misinformation category contains terms similar to 'scam' and 'conspiracy' from our dataset that helped capture references of such words in the context of COVID-19 vaccines. High reporting of adverse effects and severe symptoms in rare cases leading to death [31] becomes a significant factor in increasing vaccination hesitation. The seed words given in the Health Effects category from the VAERS database led to the formation of its vocabulary containing 'restless_sleep', 'skin_sensitivity', 'hot_flash', 'flulike_symptoms', 'complications' and more words. The semantic similarity-based approach allowed customization of categories according to our dataset while ensuring the inclusion of rather noisy words like 'feverish' and 'achiness', which cannot be found precisely in medical databases.

Inequalities based on socioeconomic status, religion, race, or demographics are standard in different countries, which can lead to inconsistencies while distributing vaccines. Inequities category encapsulated terms related to socioeconomic disparities and helped us identify its impact on other emotions. Based on inspection of our dataset of tweets, we found words like "bigotry", "underprivileged", "financial_hardship" and "institutional_racism" were occurring in a highly similar context towards vaccine distribution. Expression of inequities was found to be significantly anti-correlated with faith ($p<.05$) and contentment ($p<.01$) in India. Inaccessibility to vaccines in marginalized groups has led to lower gratification and higher anxiety among these groups [32].

We analyzed tweets from five countries around different continents to get the generalized outlook towards vaccines and how they affect the global immunization process. Suppl. Figure 2 depicts Sorrow, Rage, and Misinformation during April 2021 in the United Kingdom as the central module, with the highest PageRank. The Medicines and Healthcare products Regulatory Agency of the UK issued a new advisory during that period, concluding a possible link between COVID-19 Vaccine AstraZeneca and extremely rare, unlikely occurrences of blood clots [33]. This led to the high strength of negative emotions expressed on Twitter, along with mentions of misinformation. Suppl. Figure 3 shows the alluvial diagram of Brazil. The category of Rage, which was a relatively less important and independent module in the early months, made associations with Sorrow and Misinformation in April 2021 in Brazil. This can be attributed to the highest number of cases and deaths in that period of the pandemic in Brazil [34]. In Suppl. figure 4, we can see that faith, contentment, and vaccine rollout were relatively lower than other categories during July 2020, but later in April 2021, they formed a cluster with anticipation and gained the highest relative importance in the alluvial diagram. The reason behind this could be accredited to the announcement by the Australian government of securing an additional 20 million doses of the Pfizer-BioNTech COVID-19 vaccines overnight [35]. Australia entered into four separate agreements for the supply of COVID-19 vaccines with Pfizer, AstraZeneca, Novavax, and COVAX, which totaled up the vaccine doses to some 170 million doses, as announced by their Prime Minister.

### 4.2 Related Work
Existing literature on understanding vaccine hesitancy majorly focuses on defined questions from a part of the population belonging to a specific country[36][37][38]. While such studies done on surveys can help understand the explicit reasoning provided by the individuals, they still pose a limitation on inculcating the variation in outlooks of a larger population over a long period of time. We aim to fill these gaps by studying the important events, such as vaccine trials, highest reported deaths, or import/export of new vaccines, that fueled different populations' emotions as social media platforms are highly influential due to their comprehensive access and popularity. Our psychometric analysis considers important timestamps and a broader category of emotions to understand the Before-After change and the factors with which they associate.

Identification of psychological processes that distinguish between vaccine-hesitant and receptive groups has been carried out in recent research [39]. This helps broadcast public health advisories on social media platforms by strategically taking into account the user's perspective. Effective public health interventions encouraging the uptake of COVID-19 vaccines have benefitted from psychologically oriented approaches [40][41].

Research around understanding the general sentiments towards vaccination programs in the USA and UK through analyzing social media posts has also been conducted [42]. While their work provides an overview of positive, negative, or neutral sentiments around other important global developments affiliated with COVID-19 vaccine trials, our analysis provides a much intricate granularity in understanding the nature of emotions, temporal trends, and the influencing factors that are highest correlated. Our pipeline effectively clusters the emotions and influencing factors around important timestamps based on vaccine approval with granular details.

### 4.3 Conclusion

Our study provides research and practical implications in public policymaking and research on vaccine hesitancy. Our findings offer insights into how the different stages of a pandemic and vaccination process influence emotions and crucial factors like misinformation, health discussions, and socioeconomic disparities on Twitter. This can help decision-makers to navigate better solutions in future waves of COVID-19 or similar outbreaks and design appropriate interventions. Our approach can also be utilized to understand the general perception of people during such situations and what preventive measures should be implemented, taking the various influencing factors into account.

The future work can be attributed in the direction of local region-level analysis for a specific country to understand the granular emotions within different sections of people and the contributing factors behind it. Providing some weightage to the number of reshares and likes the social media post gets can also play an essential role in including the influence the post had in calculating overall strength. Our approach has high adaptability and can be utilized for any online forum, news, or survey data to extract various insights. Designing categories and performing temporal analysis on social media data can also be used for identifying multiple ongoing issues like the unavailability of medical resources like oxygen concentrators, ICU beds, and drugs during the second wave of COVID-19. Such analysis can be taken into account while formulating quality allocation of scarce resources based on various factors and their strength. Better information extraction and understanding of such data can be facilitated through our work.

## 5. Funding

This work was supported by the Delhi Cluster- Delhi Research Implementation and Innovation (DRIIV) Project supported by the Principal Scientific Advisor Office, Prn.SA/Delhi/Hub/2018(C).

## 6. Conflict of Interest

The authors declare that they have no known competing financial interests or personal relationships that could have appeared to influence the work reported in this paper.

## 7. Acknowledgments

This work was supported by the Delhi Cluster- Delhi Research Implementation and Innovation (DRIIV) Project supported by the Principal Scientific Advisor Office, Prn.SA/Delhi/Hub/2018(C).

# Supplementary Information

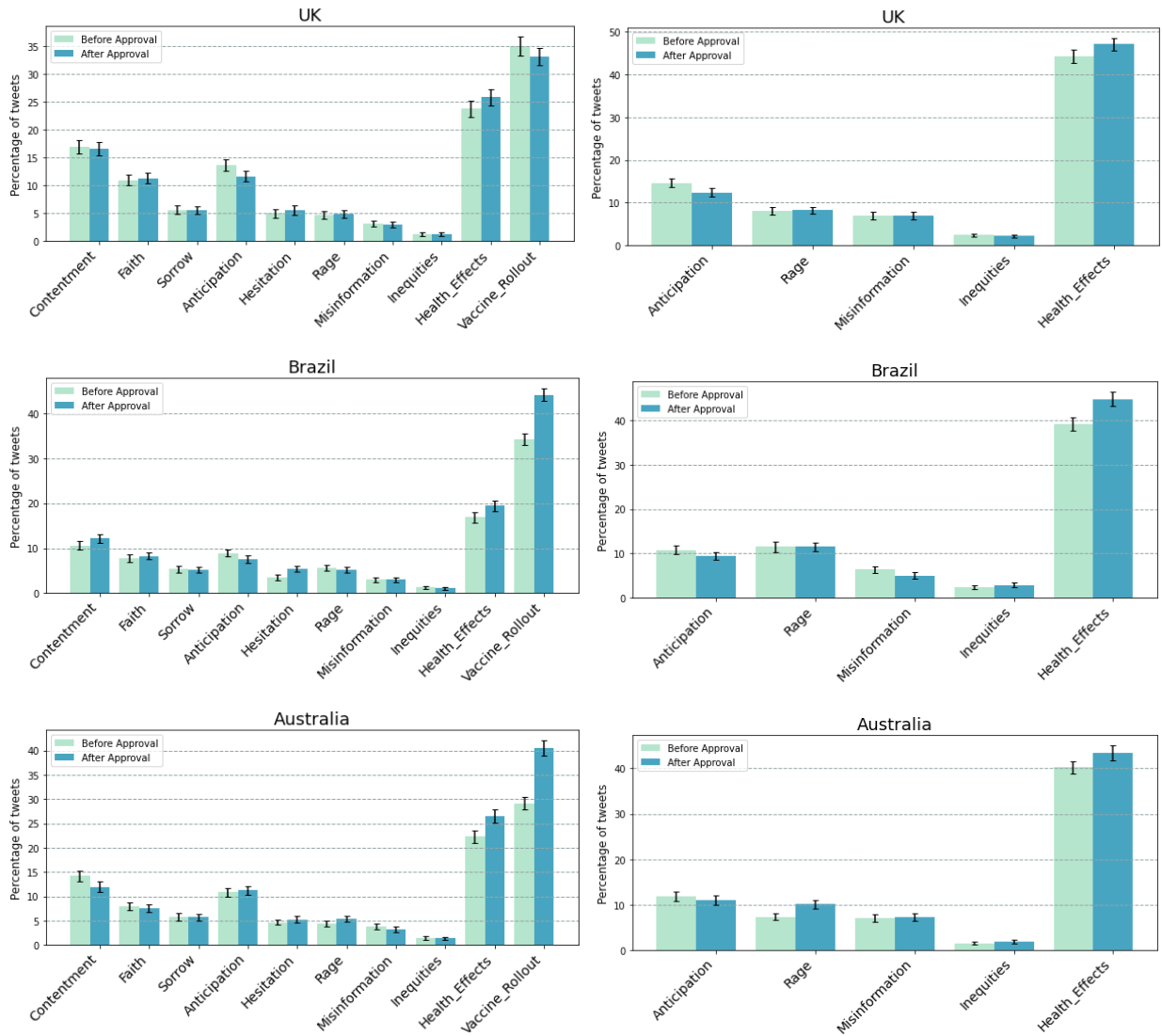

Supplementary Figure 1: (Left): Percentage of tweets having a positive strength in each lexical category before and after approval of COVID-19 vaccine in UK, Brazil and Australia. (Right): Percentage of Anticipation, Rage, Misinformation, Inequities, and Health Effects in positive 'hesitancy' tweets from UK, Brazil and Australia.

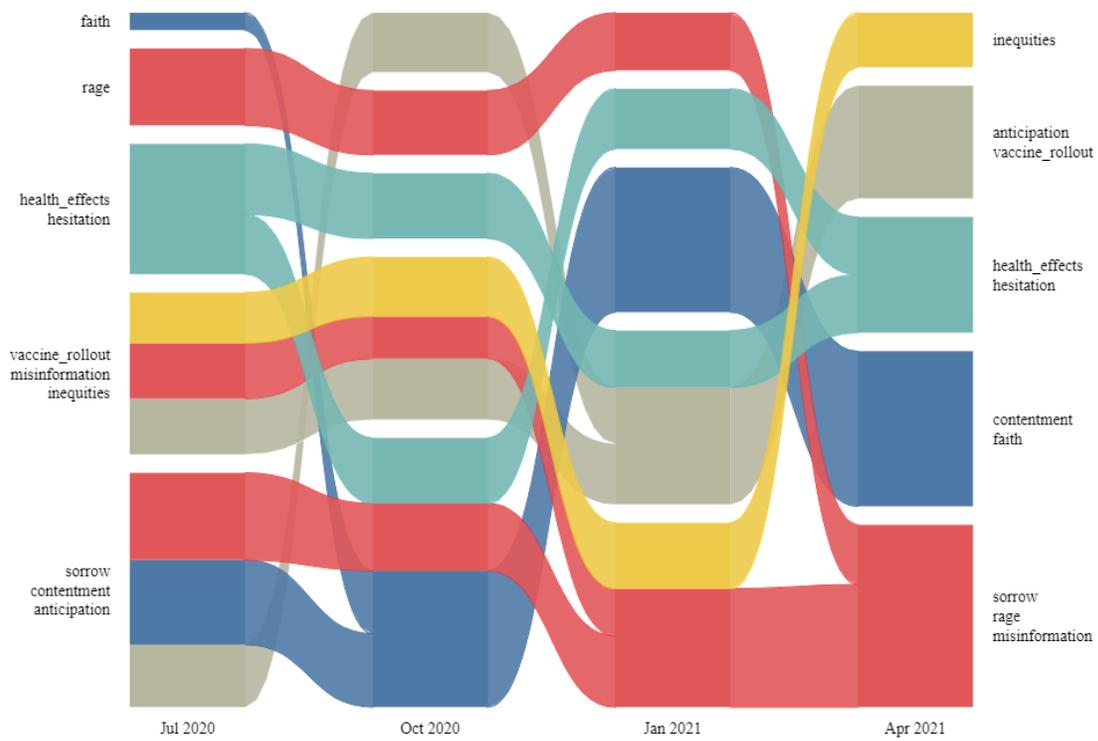

*Supplementary Figure 2: Alluvial diagram for correlation-based networks showing the evolution of categories from July 2020 to April 2021, on an interval of three months, in the UK.*

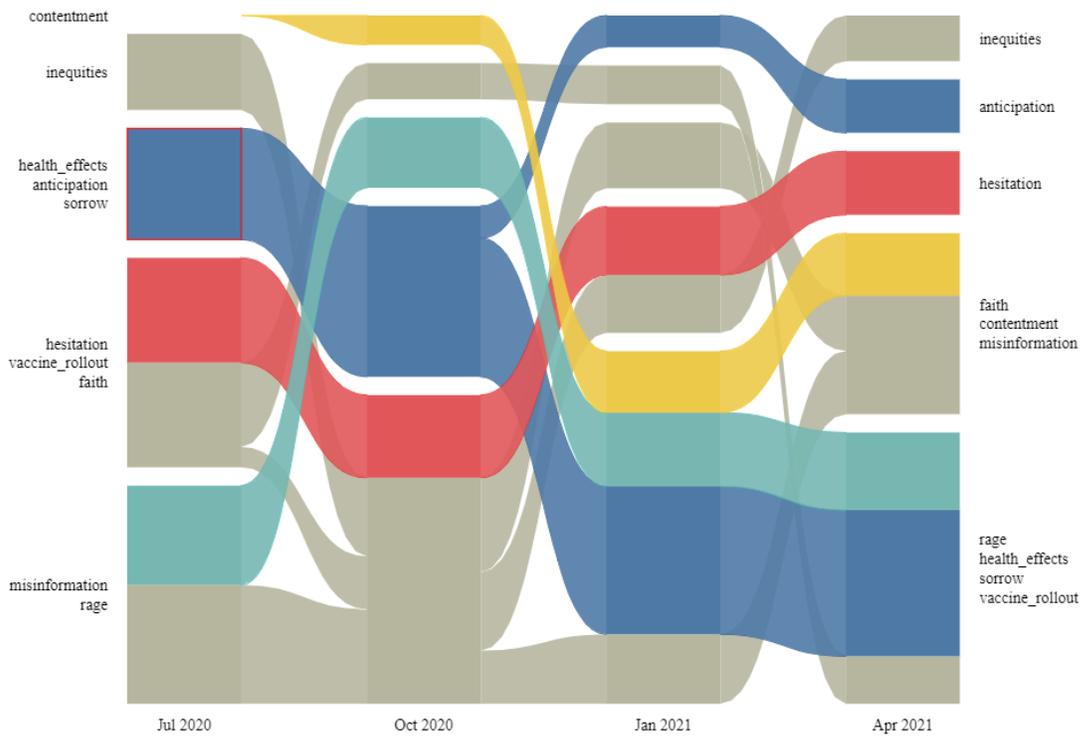

*Supplementary Figure 3: Alluvial diagram for correlation-based networks showing the evolution of categories from July 2020 to April 2021, on an interval of three months, in Brazil.*

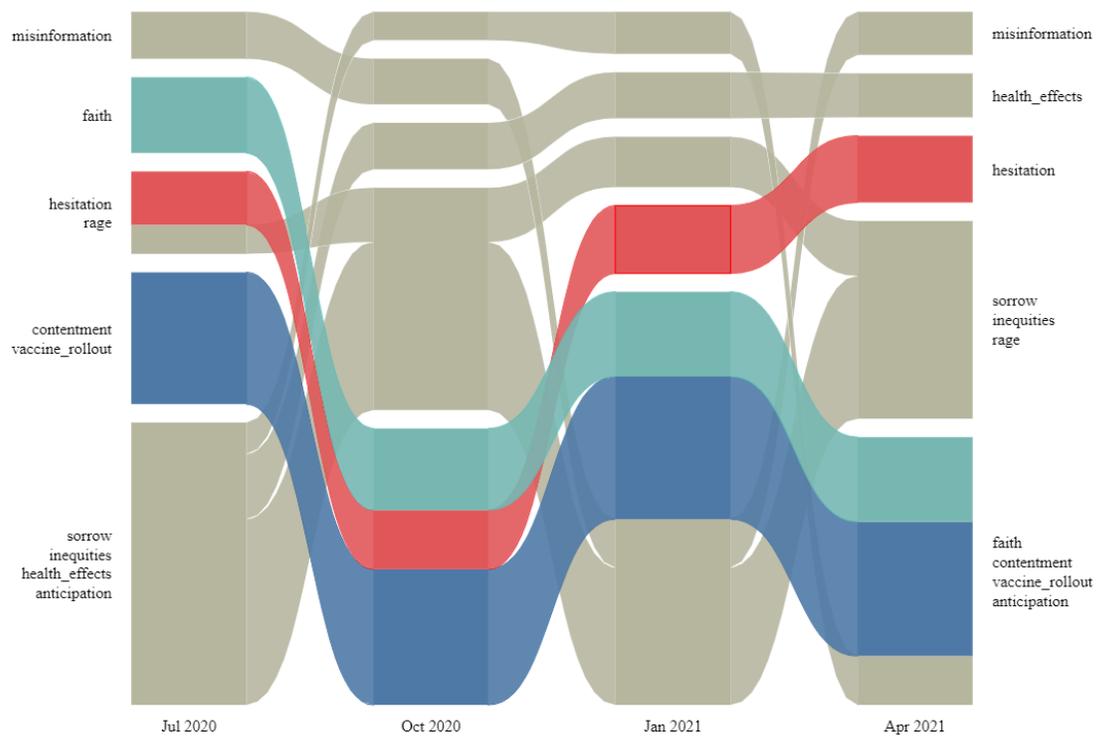

*Supplementary Figure 4: Alluvial diagram for correlation-based networks showing the evolution of categories from July 2020 to April 2021, on an interval of three months, in Australia.*